\UseRawInputEncoding

\documentclass[conference]{IEEEtran}
\IEEEoverridecommandlockouts    

\usepackage{cite}
\usepackage[pdftex]{graphicx}
\usepackage{float}
\usepackage{subfigure}
\usepackage{graphicx}
\usepackage{epstopdf}
\usepackage{multicol}
\usepackage{mathtools}
\usepackage{amsthm,amsmath,amssymb}
\usepackage{mathrsfs}
\usepackage[linesnumbered,ruled,vlined]{algorithm2e}
\usepackage{color}

\title{IA Planner: Motion Planning Using Instantaneous Analysis for Autonomous Vehicle in the Dense Dynamic Scenarios on Highways}

\author{Xiaoyu Yang and Huiyun Li$^{*}$
	\thanks{This work was supported by CAS Key Laboratory of Human-Machine Intelligence-Synergy Systems, Shenzhen Institutes of Advanced Technology, and Shenzhen Engineering Laboratory for Autonomous Driving Technology, the Science and Technology Development Fun, Macao S.A.R. (FDCT) No.0015/2019/AKP.}
	\thanks{Xiaoyu Yang and Huiyun Li are with Center for Automotive Electronics, Shenzhen Institutes of Advanced Technology, Chinese Academy of Sciences, Shenzhen 518055, China, University of Chinese Academy of Sciences, Beijing 100049, China (email: xy.yang1@siat.ac.cn; hy.li@siat.ac.cn)}%
	\thanks{$^{*}$Correspondence is {\tt\small hy.li@siat.ac.cn}}%
}

\begin{document}
\maketitle

\begin{abstract}
In dense and dynamic scenarios, planning a safe and comfortable trajectory is full of challenges when traffic participants are driving at high speed.	The classic graph search and sampling methods first perform path planning and then configure the corresponding speed, which lacks a strategy to deal with the high-speed obstacles. Decoupling optimization methods perform motion planning in the S-L and S-T domains respectively. These methods require a large free configuration space to plan the lane change trajectory. In dense dynamic scenes, it is easy to cause the failure of trajectory planning and be cut in by others, causing slow driving speed and bring safety hazards. We analyze the collision relationship in the spatio-temporal domain, and propose an instantaneous analysis model which only analyzes the collision relationship at the same time. In the model, the collision-free constraints in 3D spatio-temporal domain is projected to the 2D space domain to remove redundant constraints and reduce computational complexity. Experimental results show that our method can plan a safe and comfortable lane-changing trajectory in dense dynamic scenarios. At the same time, it improves traffic efficiency and increases ride comfort.
\end{abstract}

\IEEEpeerreviewmaketitle

\section{Introduction}
The motion planning module plays a very important role in autonomous driving technology \cite{7339478}. It needs to plan a safe and comfortable trajectory in real time for various scenarios, which largely determines the intelligence level of autonomous vehicle \cite{schwarting2018planning}. Some classic algorithms perform well in simple scenarios such as closed parks and highways with few vehicles. However, many complex scenarios are still full of challenges, and dense dynamic scenarios on highways are one of them \cite{8715479}. These scenes have the following characteristics:
\begin{itemize}
	\item The relative distance between vehicles changes rapidly, making the risk of collision difficult to model and predict;	
	\item The force analysis of ego car is complicated, making the future state difficult to estimate; 
	\item The real-time requirement of the algorithm is very high, which means that a safe and comfortable trajectory should be planned quickly and accurately.
\end{itemize}

Trajectory refers to a series of states in spatio-temporal domain.
The graph search method and the sampling method first perform a path search in the spatial domain, and then matche the corresponding speed. They have good performance in low-speed sparse scenes. However, it is difficult to dynamically describe the movement of obstacles  in the grid map in the short period of re-planning. So they are not suitable for dynamic dense scenes.
The decoupling optimization methods split motion planning into two processes in S-L domain and S-T domain (or S-T domain and L-T domain), which can reduce the complexity of the solution and improve the stability. However, such a split will cause a lot of free configuration space to be unnecessarily occupied by dynamic obstacles. Therefore, these methods usually only consider low-speed and static obstacles. And eliminate the risky trajectory, after the collision detection between planned trajectory and dynamic obstacle trajectory. 

\begin{figure}[htbp] 
	\centering
	\includegraphics[height=2.0cm,width=8.5cm,angle=0,scale=1.0]{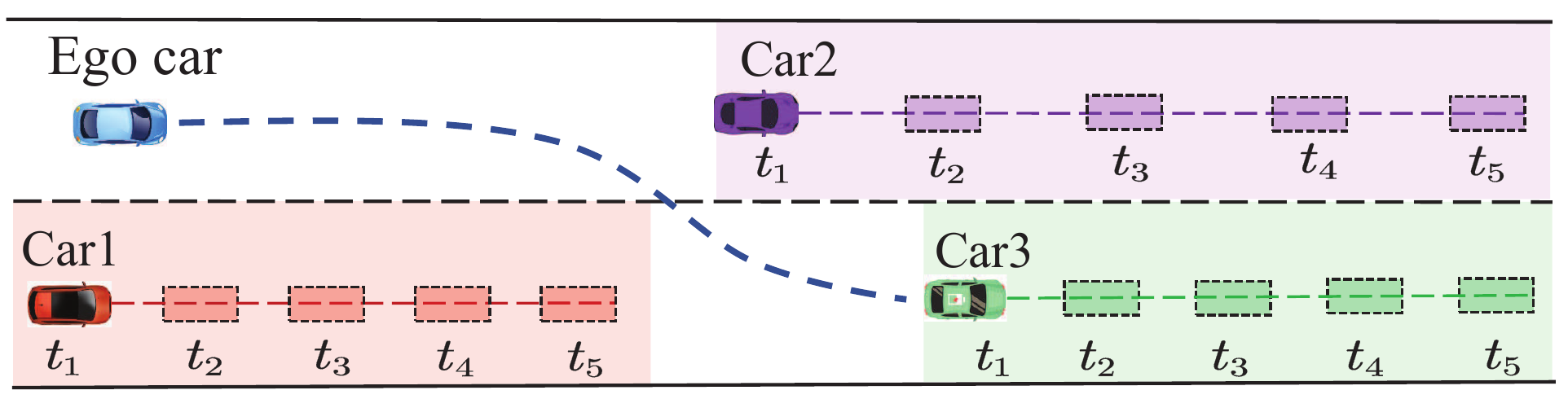} 
	\caption{ Illustration of narrow free configuration space caused by existing motion planning methods. }
	\label{fig:1} 
\end{figure} 

The shortcomings of requiring a large free configuration space to complete lane changing will easily lead to the failure of motion planning in dense dynamic scenes.
Also, as a transitional stage, autonomous vehicles and human-driven vehicles will co-drive on the road for a long time. This will cause autonomous vehicle to be cut in  easily, paralyzing the metastable high-density traffic flow \cite{2000Experimental} and causing stop-and-go waves\cite{2008Traffic}. It also causes frequent acceleration and deceleration of ego car, which increases fuel consumption and reduces safety and comfort\cite{vcivcic2019stop}.

In this article, we analyze the collision relationship in the spatio-temporal domain, and propose an instantaneous analysis model which only analyzes the collision relationship at the same time. The collision-free constraints in 3D spatio-temporal domain is projected to the 2D space domain to remove redundant constraints and reduce the difficulty of solution. Finally, the motion planning problem is modeled as a numerical optimization problem using model predictive control. In the optimization model, the state of the ego car is estimated using the vehicle dynamics model, and roll constraints and slip constraints are introduced to ensure safety.
The main contributions and innovations of this article:
\begin{itemize}
   \item The instantaneous analysis model is proposed to transform the collision relationship in the three-dimensional spatio-temporal domain into the two-dimensional space domain, removing redundant constraints;
   \item The heuristic speed is used to replace the guidance of a single target point, which improves the continuity of the trajectory and the smoothness of the longitudinal movement.
\end{itemize}

The remainder of the paper is structured as follows: In Sec. II, we review the related literature. The detailed mathematical derivation process of our motion planning method is carried out in the Sec. III and IV. In Sec. V we show results from our experiments, and in Sec. VI we discuss our conclusions.

\section{Related Works}

\subsection{Sampling-based  and Grid-based Methods}

The early motion planning methods of autonomous vehicles mainly came from research results in the field of robotics.
In the late 1990s, the proposal of PRM (Probabilistic Road Maps) and RRT (Rapidly-exploring Random Trees) opened up the field of sampling-based motion planning \cite{chi2018risk}. To solve the path planning in dynamic scenes, many variants were proposed. Wang et al. proposed EB-RRT to solve obstacle avoidance in dynamic scenes while ensuring the optimality\cite{wang2020eb}. 
Among the methods of graph search, the most representative methods are Dijkstra, A*, and state lattice algorithm. As an extension of Dijkstra, A* was deployed on Junior in the DARPA Urban Challenge and achieved good performance\cite{2008Odin}.
Similar to graph search methods, curve interpolation methods, such as spline curves\cite{piazzi2002quintic}, clothoid curves\cite{brezak2013real} and polynomial curves\cite{petrov2014modeling}, are widely used because of low computational cost.
These above methods have excellent performance in low-speed, sparse scenes. However, it is difficult to express the obstacle movement in a dynamic scene through the map.

\subsection{Overall Optimization Methods}

The trajectory optimization method is formulating the motion planning as an optimization problem. Model Predictive control (MPC) has been proven well suited for formulating the problem\cite{andersen2019trajectory}. MPC can take the updating of the environment into account during the planning process, because of its ability to handle multi-constraints\cite{schwarting2017safe}.
Among all kinds of constraints, the construction of collision-free constraints is full of challenges and can easily lead to non-convex problems\cite{rosmann2017kinodynamic}.
It is a common practice to treat obstacles as discs, but this will reduce the drivable area and cause difficulty in solving. If the obstacles are treated as completely enveloped polygons, this will make the optimization model a mixed integer programming problem \cite{da2019collision}. The dynamic programming method can solve the problem, but it is very time-consuming when there are many obstacles \cite{richards2005mixed}. Zhang proposed the Hybrid Optimization-based Collision Avoidance (H-OBCA) method to continuousize discrete variables and improve the speed of solution \cite{zhang2018autonomous}, but it is still difficult to deal with nose-to-tail traffic and at high speed scenarios \cite{mayne2000constrained} .

\subsection{Decoupling Optimization Methods}
Werling. et al. transformed the problem of structured road trajectory planning from Cartesian coordinate system to Frenet coordinate system, and completed the planning in two dimensions of longitudinal S-T domain and lateral L-T domain \cite{werling2012optimal}. The other is the path/speed decoupling method, which respectively performs path planning in the S-L domain and speed planning in the S-T domain \cite{xu2012real}. This method has better computational efficiency. Ding. et al. uses the spatio-temporal semantic corridor (SSC) method on this basis to uniformly express obstacles and traffic rules. The Apollo of Baidu has also made some improvements to improve the optimality of the trajectory through multiple rounds of path speed iteration \cite{fan2018baidu}\cite{zhou2020dl}, and the hot start method to improve the real-time performance of the algorithm \cite{he2020tdr}. Because decoupling methods can only consider two dimensions at a time, usually only static obstacles are considered in the planning process, then collision detection is performed with the predicted obstacle trajectory. So it is still difficult to deal with dense dynamic scenes.

\section{ Problem formulation }

Before introducing our method, let us make the following assumptions:
\begin{itemize}
	\item The perception system can obtain information about obstacles 300 meters ahead and 100 meters behind;
	\item Through the prediction module, the pose $ \boldsymbol{X}_{1:n}^{j} $ of the $ \boldsymbol{m} $ obstacles at the next $\boldsymbol{n}$  moments can be obtained, namely $ \left\{ x_i,y_i,v_i \right\} $, where $ i\in 1,\cdots ,n $ , $ j\in 1,\cdots ,m $. And the corresponding covariance matrix is $\boldsymbol {\varSigma} \in \mathbb{R}^{3\times 3}$, which represents the uncertainty.
\end{itemize}


We formulate a general discrete time constrained optimization with  $\boldsymbol{n}$ timesteps and time horizon $
\tau =\varSigma _{i=1}^{n}\varDelta t_i$ . We use the following notation for a set of states $X_{1:n}=\left[ X_1,\cdots ,X_n \right] \in \mathcal{Z}^n$  and for a set of inputs $U_{1:n-1}=\left[ U_i,\cdots ,U_{n-1} \right] \in \mathcal{U}^{n-1}$.

The objective is to compute the optimal inputs $U_{1:n-1}$ and state $X_{1:n}$ for the ego-vehicle which minimize a cost function $\varSigma _{i=1}^{n}J\left( X_i,U_i \right)$ to improve the comfort of the trajectory.

\begin{equation}
	\begin{aligned}
		U_{1:\boldsymbol{n}-1}^{*},X_{1:\boldsymbol{n}}^{*}=\boldsymbol{arg}\min \sum_{i=1}^n{}J\left( X_i,U_i \right)
	\end{aligned}	
\end{equation}
\begin{center}
	\begin{align}
	s.t.\ \ &X_{i+1}=f\left( X_i,U_i \right) \tag{1a}\\
		&\mathcal{B}\left( X_i \right) \cap \mathcal{S}=\oslash \tag{1b}\\
		&\mathcal{B}\left( X_i \right) \cap \mathcal{R}=\oslash \tag{1c}\\
		&\mathcal{B}\left( X_i \right) \cap \mathcal{C}=\oslash \tag{1d}\\
		&X_{\min}\leqslant \boldsymbol{X}_{\boldsymbol{i}}\leqslant X_{\max}	\tag{1e}\\
		&U_{\min}\leqslant \boldsymbol{U}_i\leqslant \,\,U_{\max} \tag{1f}\\
		&\forall i\in \left\{ 1,\cdots ,n \right\} 	\tag{1g}\\
		&\forall j\in \left\{ 1,\cdots ,m \right\}    \tag{1h}
	\end{align}
\end{center}

The optimization is subject to a set of constraints, which represent:
\begin{itemize}
	\item the transition model of the ego-vehicle  $X_{i+1}=f\left( X_i,U_i \right)$	
	\item no slip constraint $\mathcal{B}\left( X_i \right) \cap \mathcal{S}=\oslash$	
	\item no rollover constraint  $\mathcal{B}\left( X \right) \cap \mathcal{R}=\oslash$
	\item collision-free constraint  $\mathcal{B}\left( X \right) \cap \mathcal{C}=\oslash$, we will explain in detail in Sec. VI.	
	\item the upper and lower bound conditions $X_{\min}\leq X_i\leq X_{\max}, U_{\min}\leq U_i\leq \,\,U_{\max}$ that the vehicle state quantity and control quantity need to meet.
\end{itemize}

Given the estimated trajectories $\left( \boldsymbol{X}_{1:n}^{j},\boldsymbol{\varSigma }_{1:n}^{j} \right)$ for all traffic participants $j=1,\cdots ,m$ and the initial state $\boldsymbol{X}_0$ of the ego vehicle, the optimal trajectory for the ego-vehicle is then given by the following receding horizon optimization,

\subsection{ Cost function }

We punish the longitudinal acceleration   $\boldsymbol{a}$ and the steering angular velocity of the front wheels $\boldsymbol{\gamma }$  to ensure that the resulting trajectory is safe and smooth. $\boldsymbol{w}_1$ and $\boldsymbol{w}_2$ respectively represent the weight of each penalty item. The third part of the cost function is used to punish the error of the velocity $v$ and the heuristic value ${{v}_{pre}}_i$. The weight $w_3$ is used to adjust the closeness of them.

\begin{equation}
	J\left( X_i,U_j \right) =w_1\cdot a_{i}^{2}+w_2\cdot \gamma _{i}^{2}+w_3\cdot \left( v-v_{pre_i} \right) ^2 
\end{equation}

\begin{figure}[htbp] 
	\centering
	\subfigure[X-T domain]{
		\begin{minipage}[t]{0.25\linewidth}
			\centering
			\includegraphics[height=5cm,width=5.5cm,angle=0,scale=1.0]{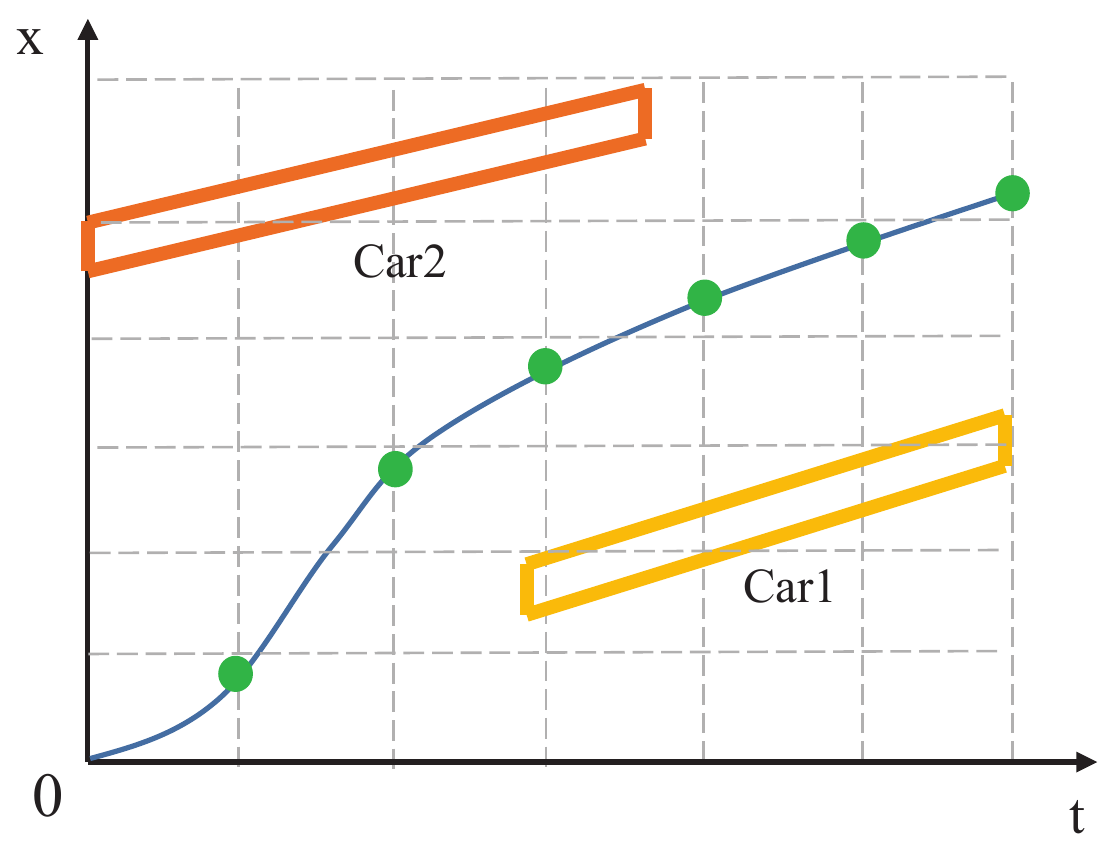}
		\end{minipage}
	}
	\hspace{3cm}
	\subfigure[X-Y domain]{
		\begin{minipage}[t]{0.25\linewidth}
			\centering
			\includegraphics[height=5cm,width=2.5cm,angle=0,scale=1.0]{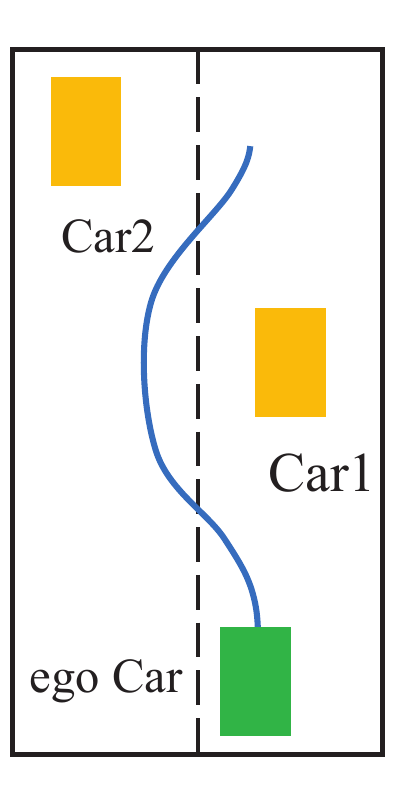}
		\end{minipage}
	}	
	\caption{ Illustration of longitudinal heuristic speed planning. This step just gets the heuristic value as $\boldsymbol{\mathrm{v}_{pre}}_{1:n}$, which will be adjusted in the overall optimization model. }
	\label{fig:2} 
\end{figure}

 The heuristic value ${{v}_{pre}}_i$ can be obtained in advance through speed planning in the $X-T$ domain. First, the search space is discretized, and the obstacle information is projected into the $X-T$ domain. Then, the Hybrid A* method can be used to quickly search for the curve required for overtaking or lane changing. Finally, the slope of the curve $X-T$ can approximate the heuristic value $v_{pre}$ at each moment. It should be noted that these are only initial heuristic values and will be adjusted in the overall optimization. This methods can replace the guiding role of the target point, while avoiding the complexity of target point selection and increasing the continuity of longitudinal motion.

\subsection{ Vehicle dynamic model }

\begin{figure}[htbp] 
	\centering
	\includegraphics[height=5.8cm,width=8cm,angle=0,scale=1.0]{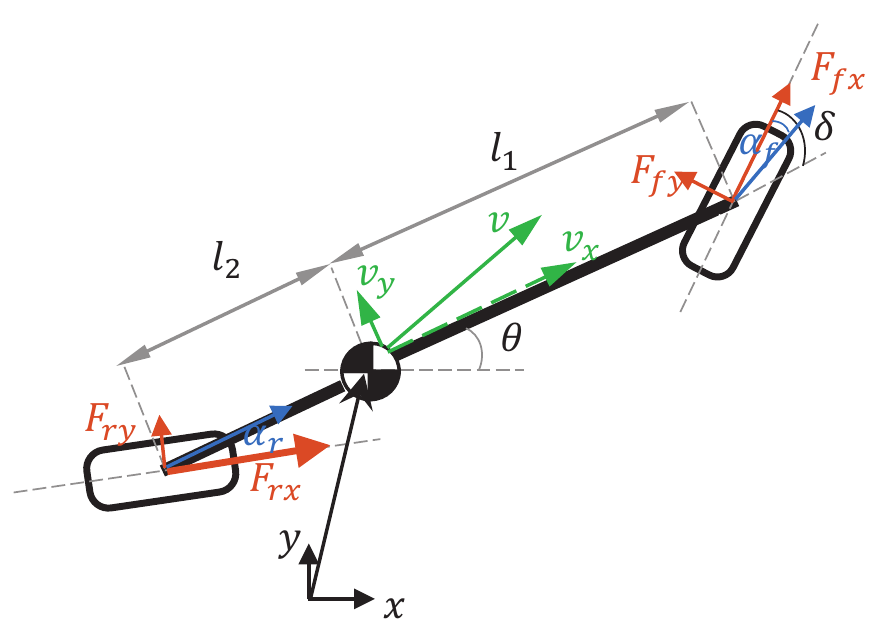} 
	\caption{ Dynamic bicycle model of the Ackermann-steered vehicle\cite{schwarting2017safe}.} 
	\label{fig:3} 
\end{figure} 

The state variables are
\begin{itemize}
	\item $x,y,\theta$: the pose of ego car at rear differential	
	\item $v_x,v_y$: longitudinal/lateral velocity at center of gravity
	\item $r$: yaw rate
\end{itemize}

The control variables are
\begin{itemize}	
	\item $\gamma$: rate of the steering angle
	\item $a$: longitudinal acceleration along the heading
\end{itemize}

The parametric variables are
\begin{itemize}
	\item $C_{\alpha f},C_{\alpha r}$: front and rear tire cornering stiffness
	\item $l_1,l_2$: distance from the center of gravity to front/rear axles
	\item $m$: vehicle mass
	\item $I_z$: yaw moment of inertia of the vehicle
\end{itemize}

\begin{figure*}[htbp]
	\begin{equation}
		\left[ \begin{array}{c}
			\dot{x}\\
			\dot{y}\\
			\dot{\theta}\\
			\delta\\
			\dot{v}_x\\
			\dot{v}_y\\
			\dot{r}\\
		\end{array} \right] 
		=
		\left[ \begin{array}{c c c c c c c}
			0&	0&	0 & 0&		1&		0&		0\\
			0&	0&	0 & 0&		0&		1&		0\\
			0&	0&	0 & 0&		0&		0&		1\\
			0&	0&	0 & 0&		0&		0&		0\\
			0&	0&	0 & \frac{2C_{\sigma f}\delta}{m}&	0&	   \frac{2C_{\sigma f}\delta}{mv_x}&		v_y-\frac{2C_{\sigma f}a\delta}{mv_x}\\
			0&	0&	0 & \frac{2C_{\alpha f}}{m}&		   0&		\frac{-2C_{\alpha f}-2C_{\alpha r}}{mv_x}&		-v_x+\frac{2C_{\alpha r}b-2C_{\alpha f}l_1}{mv_x}\\
			0&	0&	0 & \frac{2C_{\alpha f}a}{I_z}&		0&		\frac{2C_{\alpha f}b-2C_{\alpha f}a}{I_zv_x}&		\frac{-2C_{\alpha f}a_{1}^{2}-2C_{\alpha r}b_{2}^{2}}{I_zv_x}\\
		\end{array} \right] 
		\cdot 
		\left[ \begin{array}{c}
			x\\
			y\\
			\theta\\
			\delta\\
			v_x\\
			v_y\\
			r\\
		\end{array} \right]
		+\left[ \begin{array}{c c}
			0&		0\\
			0&		0\\
			0&		0\\
			1&		0\\
			0&		1\\
			0&		0\\
			0&		0\\
		\end{array} \right] 
		\cdot 
		\left[ \begin{array}{l}
			\gamma\\
			a\\
		\end{array} \right] 
	\end{equation}
\end{figure*}

The vehicle dynamics model in continuous space is shown in equation 3, where the selected state quantity $X=\left[ x,y,\theta ,\delta ,v_x,v_y,r \right] ^T$ is  and the control quantity is $U=\left[ \gamma ,a \right] ^T$. Therefore Equation 3 can be sorted into the following Equation 4:
\begin{equation}
	X_{i+1}=f\left( X_i,U_i \right)
\end{equation}
The state equation in discrete form can be approximated as,
\begin{equation}
	X_{i+1}=f\left( X_i,U_i \right) =X_i+\int_i^{i+\varDelta t}{\dot{X}}dt
\end{equation}

It is important to get a discrete version of the vehicle dynamics by integrating the equation over a short period of time. To get more accurate forward prediction over a short period of time, we can also obtain the discrete-time model by performing numerical integration with various methods, e.g., the Runge–Kutta 4-th order method:
\begin{equation}
	X_{i+1}=f\left( X_i,U_i \right) =X_i+\frac{\varDelta t_i}{6}\left( k_1+k_2+k_3+k_4 \right)
\end{equation}
where
\begin{equation}
	\begin{aligned}
		k_1&=f\left( X_i,U_i \right)                       \\
		k_2&=f\left( X_i+\frac{dt}{2}\cdot k_1,U_i \right) \\
		k_3&=f\left( X_i+\frac{dt}{2}\cdot k_2,U_i \right) \\
		k_4&=f\left( X_i+dt\cdot k_3,U_i \right)           
	\end{aligned}
\end{equation}

\subsection{ Security constraints }
\begin{figure}[htbp] 
	\centering
	\subfigure[Roll constraint]{
		\begin{minipage}[t]{0.25\linewidth}
			\centering
			\includegraphics[height=4cm,width=4.5cm,angle=0,scale=1.0]{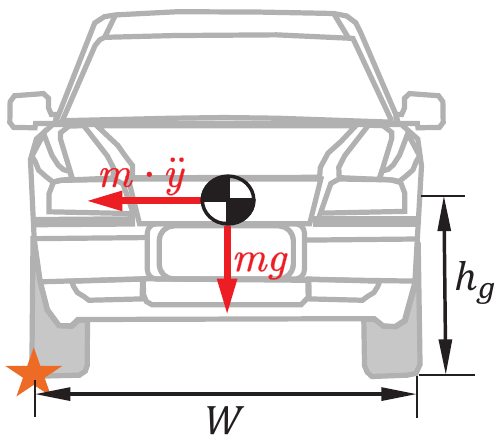}
		\end{minipage}
		}
	\hspace{2.5cm}
	\subfigure[Slip constraint]{
		\begin{minipage}[t]{0.25\linewidth}
			\centering
			\includegraphics[height=4cm,width=3cm,angle=0,scale=1.0]{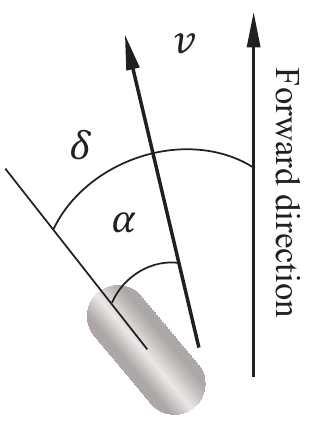}
		\end{minipage}
		}	
	\caption{ Illustration of roll and slip constraints. (a) The ratio of friction torque and gravity torque will affect the roll of the vehicle; (b) $\alpha$ represents the slip angle of the wheel. }
	\label{fig:4} 
\end{figure} 

The condition to avoid a vehicle rollover is
\begin{equation}
	\ddot{y}_i\leqslant \frac{W\cdot g}{2h_g}\times \eta 
\end{equation}
$\ddot{y}_i$ can be obtained through lateral force analysis based on Newton's second theorem. That is
\begin{equation}
	m\ddot{y}_i=-mv_{x_i}\ddot{\theta}+2F_{fy}+2F_{ry}, i\in \left[ 1,N \right]   
\end{equation}
Further calculation can get
\begin{equation}
		\ddot{y}_i=-v_{x_i}\dot{\theta}+2\left[ \frac{C_{\alpha f}}{m}\left( \delta _i-\frac{v_{y_i}+l_1r_i}{v_{x_i}} 	\right) +C_{\alpha r}\frac{l_2r_i-v_{y_i}}{mv_{x_i}} \right]
		$$
		$$
		i\in \left[ 1,n \right] 
\end{equation}

In order to prevent the vehicle from slipping, the slip angle of the front and rear wheels must meet the conditions
\begin{equation}
	\left| \frac{v_{y_i}+l_1r}{v_{x_i}}-\delta _i \right|\leqslant \alpha _{f\_\max}\text{，}
	$$
	$$
	\,\,  \left| \frac{v_{y_i}-l_2r}{v_{x_i}} \right|\leqslant \alpha _{r\_\max}\text{，}
	$$
	$$
	\,\,   i\in \left[ 1,n \right] 
\end{equation}

\section{Collision-free constraint}
This section mainly explains the collision -freee constraints. In subsection A, the collision relationship at a single moment is analyzed. In subsection B, the instantaneous analysis model is explained about the collision relationship at all moments.

\subsection{Collision-free constraint at a single moment}
\begin{figure}[htbp] 
	\centering
	\includegraphics[height=5.5cm,width=6cm,angle=0,scale=1.0]{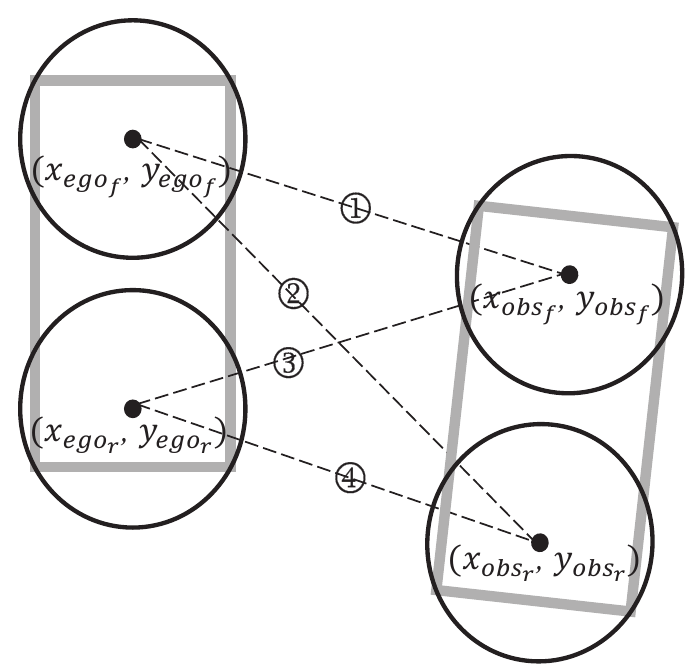} 
	\caption{ Illustration of collision-free constraints at the single moment. }
	\label{fig:5} 
\end{figure}
As shown in Figure 5, the collision-free constraint at a single moment can be expressed by Equation 12.
\begin{equation}
d_{i\_1}^{2}=\left( x_{ego_f}-x_{obs_f} \right) ^2+\left( y_{ego_f}-y_{obs_f} \right) ^2\geqslant 2R
$$
$$
d_{i\_2}^{2}=\left( x_{ego_f}-x_{obsr} \right) ^2+\left( y_{ego_f}-y_{obs_r} \right) ^2\geqslant 2R
$$
$$
d_{i\_3}^{2}=\left( x_{ego_r}-x_{obs_f} \right) ^2+\left( y_{ego_r}-y_{obs_f} \right) ^2\geqslant 2R
$$
$$
d_{i\_4}^{2}=\left( x_{ego_r}-x_{obsr} \right) ^2+\left( y_{ego_r}-y_{obs_r} \right) ^2\geqslant 2R
\end{equation}
where
\begin{equation}
x_{ego_f}=x_i+l_1\cdot \cos \theta 
$$
$$
y_{ego_f}=y_i+l_1\cdot \sin \theta 
$$
$$
x_{ego_r}=x_i-l_2\cdot \cos \theta 
$$
$$
y_{ego_r}=y_i-l_2\cdot \sin \theta 
$$
$$
R=\left( l_1+l_2 \right) /2
\end{equation}

\subsection{Instantaneous Analysis Model}

\begin{figure}[H] 
	\centering
	\includegraphics[height=6cm,width=8.5cm,angle=0,scale=1.0]{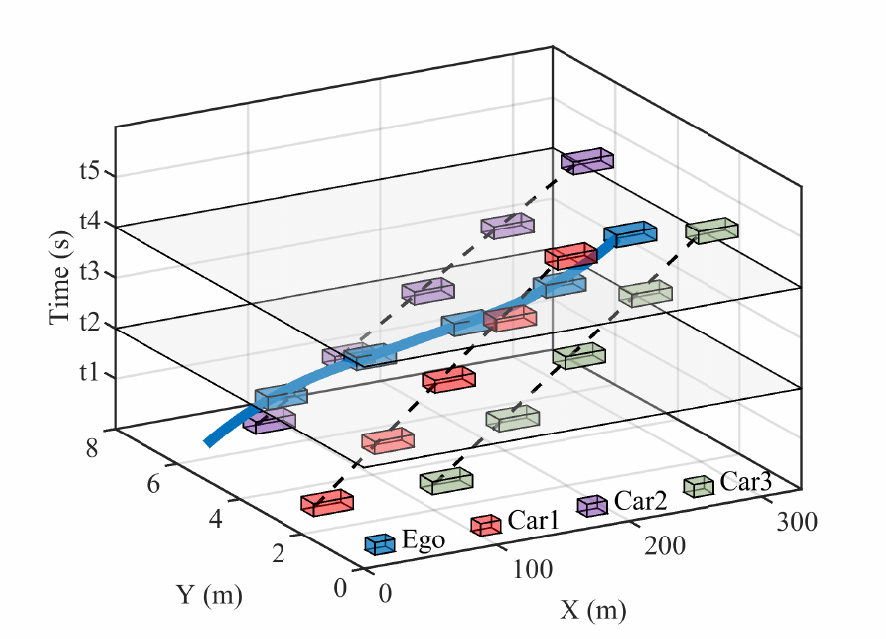} 
	\caption{ Collision relationship in 3D spatio-temporal domain. }
	\label{fig:6} 
\end{figure}

\begin{figure}[H] 
	\centering
	\subfigure[Time = $t2$]{
		\begin{minipage}[b]{0.5\textwidth}
			\centering
			\includegraphics[height=2.5cm,width=8.5cm,angle=0,scale=1.0]{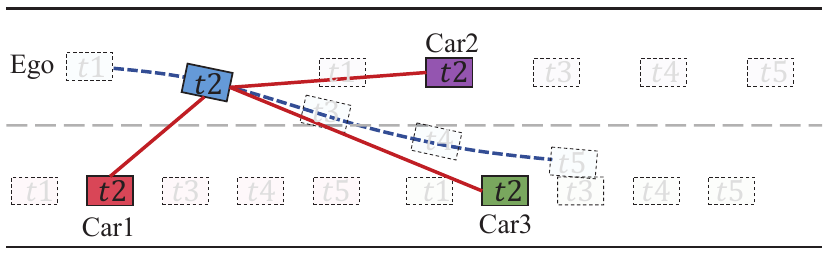} 
		\end{minipage}
	}
	\subfigure[Time = $t4$]{
		\begin{minipage}[b]{0.5\textwidth}
			\centering
			\includegraphics[height=2.5cm,width=8.5cm,angle=0,scale=1.0]{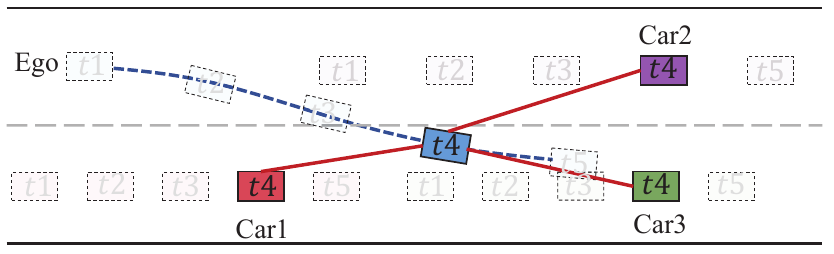}
		\end{minipage}
	}	
	\caption{ Instantaneous Analysis Model. Time is unique, so the vehicle at this time cannot collide with obstacles in the past or in the future. In figure a, the red straight line represents the collision relationship at $t2$. It can be seen that there is still ample free space. Figure b is the same. }
	\label{fig:7} 
\end{figure}

If we ignore the time dimension and only perform path planning in the $X-Y$ domain shown in Figure 1, it will cause planning failures in dense scenarios. As shown in Figure 6, after increasing the consideration of the time domain, there is sufficient free configuration space even in crowded dynamic scenes. Based on this analysis, we propose the instantaneous analysis model. When analyzing the collision relationship, we only consider the state of the ego car and the obstacle car at the same time, as shown in Figure 7. Because the ego car at this time will definitely not collide with the obstacle cars in the past or in the future.
If the state at this moment of ego car needs to be analyzed for the collision relationship with the estimated obstacle states at all times , the number of constraint-free collision conditions will be $4N^2$. 
There are a large number of constraints, and it will severely reduce the free configuration space. Eventually it will lead to the failure of trajectory planning.

\section{ Experimental results }

\subsection{ Implementation Details }
In order to verify the validity, feasibility and real-time performance of the proposed motion planning algorithm, we have built a simulation platform. The specific details are shown in Figure 8. This simulation environment runs on an Ubuntu 16.04 operating system, equipped with Intel i5-1021 CPU and 16G RAM. The solution library used is based on IPOPT （(Interior Point Optimizer). The input of  the Carla simulator is the rate of the steering angle and the longitudinal acceleration of ego car, namely $U=\left[ \gamma ,a \right] ^T$. The output is the state vector $X=\left[x,y,\theta ,\delta,v_x,v_y,r \right] ^T$ of ego car and the state information $X_{1}^{j}$ of the surrounding vehicles. We use Gaussian process to estimate the subsequent state $X_{2:n}^{j}$ of surrounding vehicles and the corresponding covariance matrix $\varSigma _{1:n}^{j}$, which is equivalent to the prediction module. Similarly, we can get the boundary of road from HD map. After the control module receives the trajectory, we use an LQR controller to complete the lateral trajectory tracking of the vehicle, and use the dual-loop PID controller to complete the longitudinal control of the vehicle.

\begin{figure}[htbp] 
	\centering
	\includegraphics[height=5.8cm,width=8.5cm,angle=0,scale=1.0]{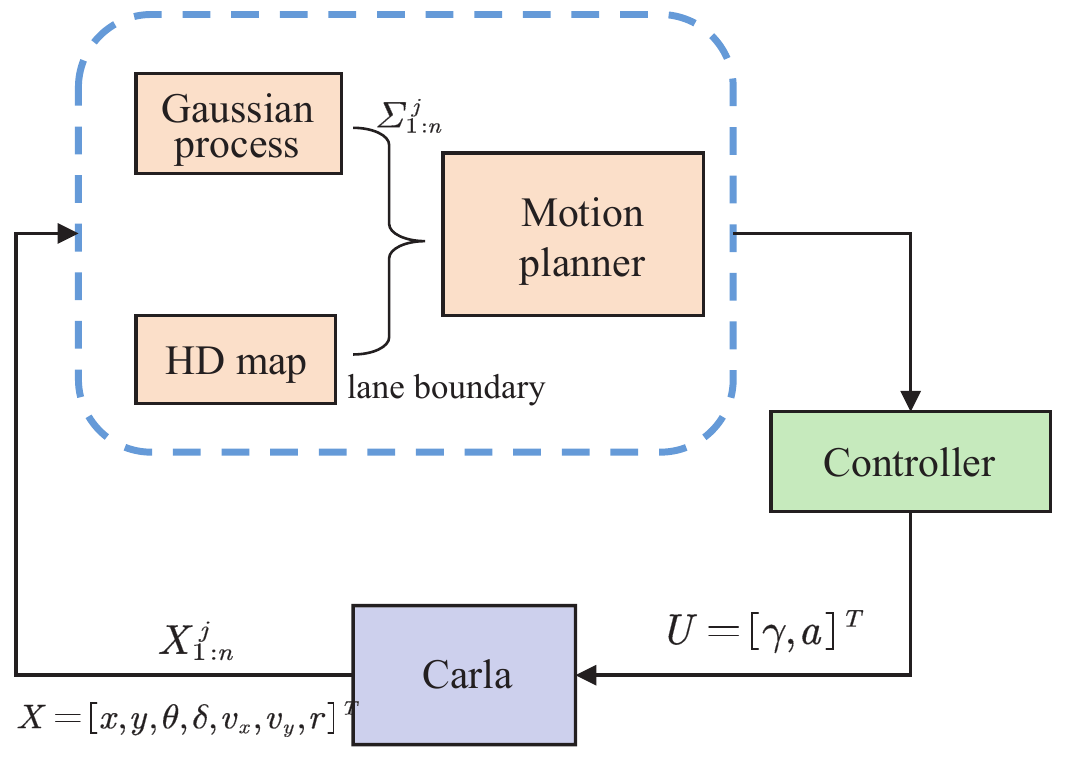} 
	\caption{ Framework of the simulation platform. } 
	\label{fig:8} 
\end{figure}

In Sec III and Sec IV, there are many physical parameters of the vehicle dynamics model and hyperparameters of the optimization model. They are displayed in Table 1.

\begin{table}[H]
	\begin{center}
		\caption{The value of each parameter}
		\begin{tabular}{c|c|c|c}
			\textbf{Parameter} & \textbf{Value} & \textbf{Parameter}& \textbf{Value} \\
			\hline 
			$C_{\alpha f}\left( N/rad \right)$ & 250000 & $w_1$                                    &  50    \\
			$C_{\alpha r}\left( N/rad \right)$ & 250000 & $w_2$                                    &  100   \\
			$l_1 \left( mm \right)$            & 1180   & $w_3$                                    &  10    \\
			$l_2 \left( mm \right)$            & 1770   & $\alpha _{f\_\max}\left( rad \right)$    &  0.052 \\
			$h_g \left( mm \right)$            & 720    & $\alpha _{r\_\max}\left( rad \right)$    &  0.052 \\
			$W   \left( mm \right)$            & 1875   & $R\left( m \right)$                      &  1.3   \\
			$m   \left( kg \right)$            & 1590   & $a_{max} \left( m/s^2 \right)$           &  2     \\
			$I_z \left( kg\cdot m^2 \right)$   & 2687   & $\gamma_{max} \left( rad/s \right)$      & $\pi/2$\\
			$\varDelta t_{1:n}\left(s\right)$  & 0.5    & $\eta$                                   &  0.5   \\
			$n$                                & 20
		\end{tabular}
	\end{center}
\end{table}

\subsection{Trajectory planned at a single moment in typical scenarios}
\begin{figure*}[htbp] 
	\centering
	\includegraphics[height=5cm,width=18cm,angle=0,scale=1.0]{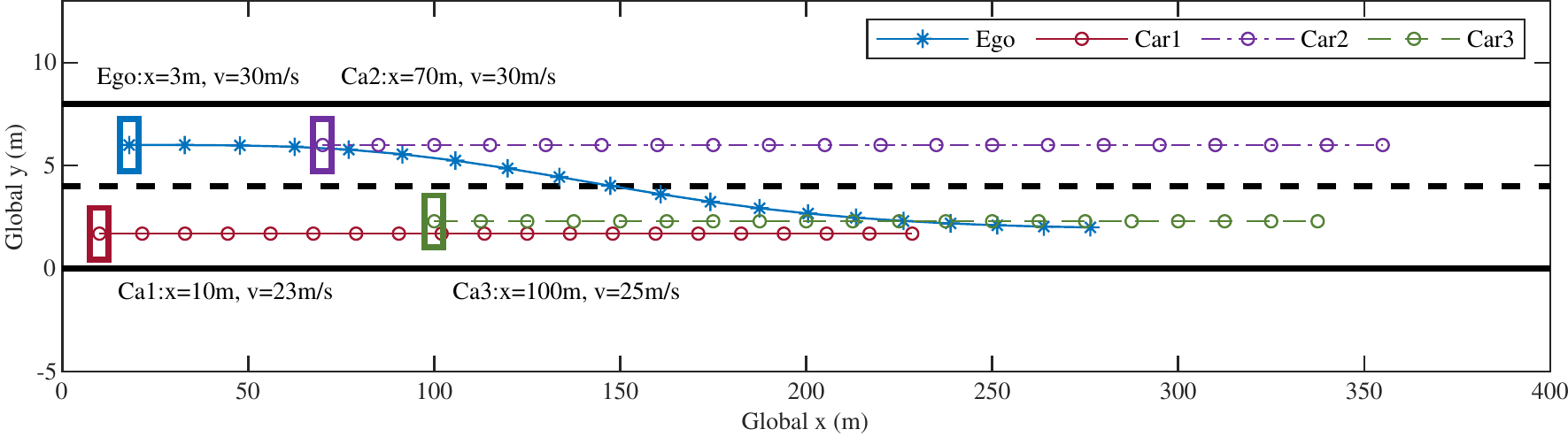} 
	\caption{ Dynamic dense scene at a single moment. The blue solid  line indicates the planned path of ego car. The circles and asterisks represent the position of each vehicle at different times.} 
	\label{fig:9} 
\end{figure*} 
\begin{figure}[htbp] 
	\centering
	\includegraphics[height=5cm,width=7.5cm,angle=0,scale=1.0]{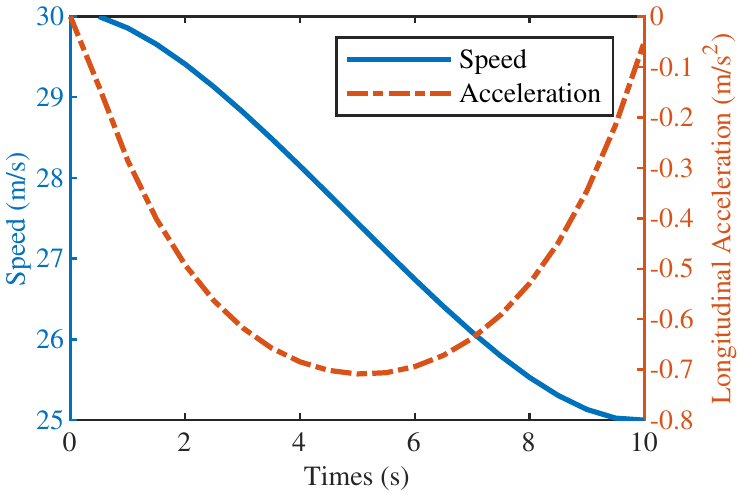} 
	\caption{The speed and acceleration of ego car corresponding to the path in figure 9.} 
	\label{fig:10} 
\end{figure} 
\begin{figure}[htbp] 
	\centering
	\includegraphics[height=6cm,width=8cm,angle=0,scale=1.0]{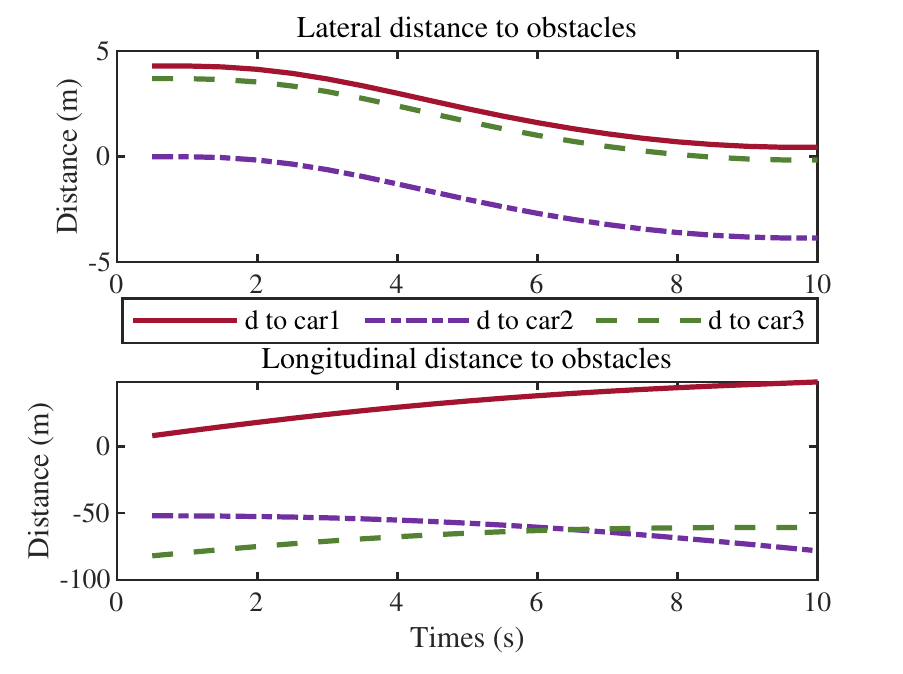} 
	\caption{ Lateral and longitudinal distance to obstacles in figure 9.  } 
	\label{fig:11} 
\end{figure}
In a dense dynamic environment, it is difficult to plan a complete overtaking trajectory at one time. In most cases, it needs to be divided into two steps: left lane change and right lane change, and there is usually a car following mode between them. In order to verify the effectiveness of the proposed method, we select a typical right lane change scene, that is, the vehicle changes from the left lane to the right lane. Although the left side is a high-speed lane, the right side is a low-speed lane. When the traffic is sparse, choosing the left lane as much as possible can naturally improve the traffic efficiency, but in dense scenarios, it may appear that the front vehicle on the left lane is slow and needs to be switched to the right lane. If we need to exit the highway ahead, we also need to switch to the right lane in advance.

\textbf{Scenario introduction}: (the initial state of each car): As shown in Figure 9, the initial state of the ego car is $\left( x_{ego}=20,y_{ego}=6,\theta _{ego}=0^{\circ},v_{ego}=30m/s,\delta _{ego}=0^{\circ} \right)$, and the initial acceleration is $\left( a_{ego}=0m/s^2 \right)$. The initial state of car1 is$\left( x_{car1}=5m,y_{car1}=1.8m,v_{car1}=23m/s \right)$. The initial state of car2 is $\left( x_{car2}=100m,y_{car2}=2.2m,v_{car2}=25m/s \right)$. The initial state of car3 is $\left( x_{car3}=70m,y_{car3}=60m,v_{car3}=30m/s \right)$.

\textbf{During lane changing}: Figure 11 shows the lateral and longitudinal distance between the vehicle and the other three vehicles during the lane change. In the first $5$ seconds, the ego car is still in the left lane. The lateral distance from car2 is almost $0$ meters, and the minimum longitudinal distance is $50m$. So there is no collision with car2. From $5s$ to $10s$, when the ego vehicle enters the right lane, the lateral distance between it and car1 continuously decreases to $0m$. Also, the longitudinal distance from car1 is increasing to $50m$, and the longitudinal distance from car3 is reduced from $90m$ to $50m$. Therefore, there is no collision during the lane changing.

\textbf{End state}: As shown in Figure 10, the speed of the ego car is $24.5m/s$ at $10s$, between $v_{car1}=23m/s$ and $v_{car1}=25m/s$, and at this time $a_{car1}=0.05m/s^2$ is close to $0$. Ego car can enter the following mode very smoothly without emergency braking.

\textbf{Enforceability}: During lane changing, the longitudinal acceleration of the vehicle changes continuously and gently. Therefore, there is no step response for the control module, and trajectory tracking can be completed well.

\subsection{Comparative experiment in dense traffic flow}

\begin{figure}[H] 
	\centering
	\includegraphics[height=4cm,width=8.5cm,angle=0,scale=1.0]{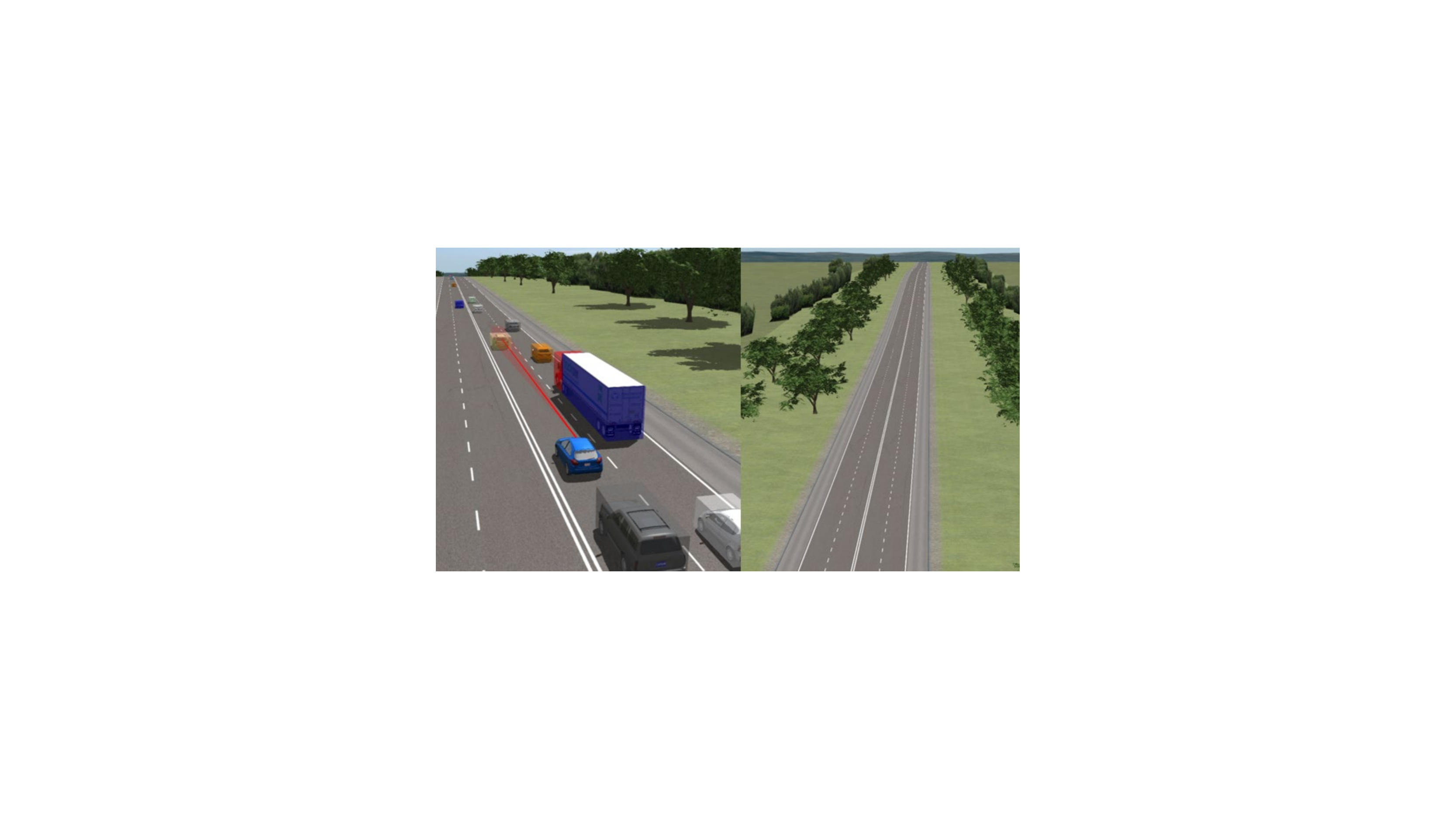} 
	\caption{ Virtual scene built in simulation experiment } 
	\label{fig:12} 
\end{figure} 

In order to verify the performance of our proposed method in a dynamic dense environment, we tested it on a road with a length of $3000m$. 
And, we make a quantitative comparison with the  seminal work \cite{werling2012optimal}. In \cite{werling2012optimal}, the S-L-T domain is discretized with a certain resolution based on the Frenet framework, and the optimal trajectory is selected according to the strategy for different behavior decisions.

We built the road scene in Roadrunner and saved it in the Opendrive format that Carla can use.
There are $20$ vehicles randomly arranged in the left lane, and the initial position spacing is from 50$m$ to 150$m$. Their initial speed is between $70km/h$ and $100km/h$ randomly, and then they travel at a constant speed.
There are $20$ vehicles randomly arranged in the right lane, and their initial position spacing is from $50m$ to $150m$. The initial speed is from $90km/h$ to $120km/h$, and then they travels at a constant speed.
During driving, the distance between some vehicles may be reduced or increased. Traffic participants which are less than 50 meters away from the front vehicle  will enter the following mode.

In the comparative experiment, we use Multipolicy Decision-Making(MPDP) \cite{galceran2015multipolicy} as the behavioral layer, which models the behavior planning problem as a partially observable Markov decision process (POMDP) to deal with the uncertainty in the dynamic environment.

\begin{figure}[htbp] 
	\centering
	\includegraphics[height=4.5cm,width=8.5cm,angle=0,scale=1.0]{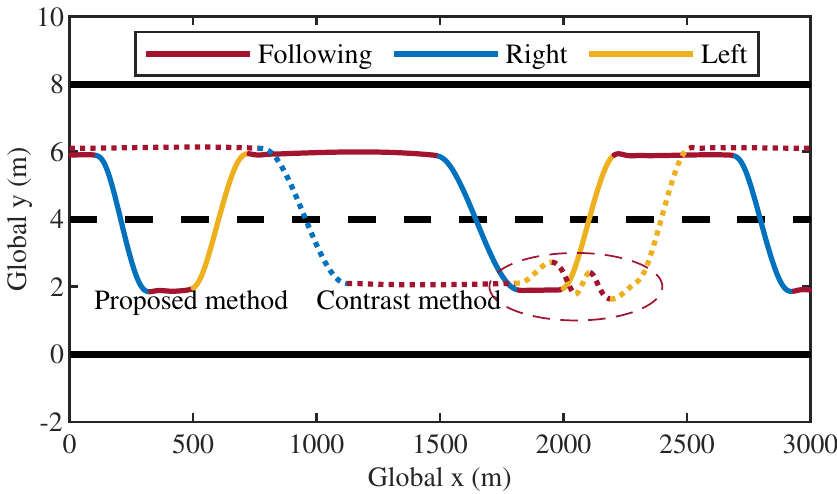} 
	\caption{ The trajectories of the two methods in the scene of Figure 12. 'Fellowing' means following a car, 'Right' means the right lane change, and 'Left' means the left lane change. The proposed method has completed five lane changes in the dense dynamic scenes. In contrast, the comparative method only completed two lane changes, and carried out frequent state switching between lane changing and following around 2000m. } 
	\label{fig:13} 
\end{figure}

It can be seen from Figure 13 that in the dense scene with a length of 3000m, our proposed method performs 5 necessary lane changes to increase the speed. In contrast, the contrast method only performed two lane changes. The reason is that this method requires a larger free configuration space to complete the lane change. Unfortunately, this is difficult to appear in a crowded traffic scene, and the opportunity  is fleeting. Moreover, this also caused frequent decision-making jumps around 2000m.

\begin{figure}[htbp] 
	\centering
	\includegraphics[height=5cm,width=8.5cm,angle=0,scale=1.0]{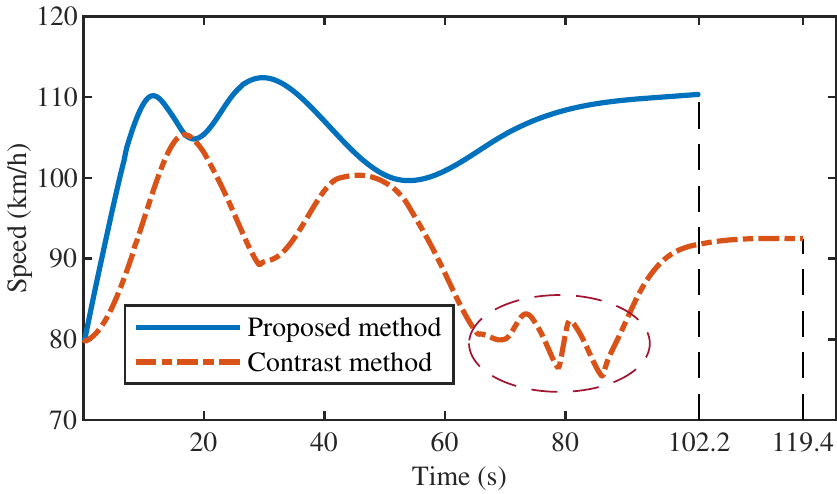} 
	\caption{ Illustration of speed curves for different methods in the scene of Figure 12. } 
	\label{fig:14} 
\end{figure}

As shown in Figure 14, compared with the contrast method, our proposed method enables the vehicle to reach the end point 17.2 seconds earlier and pass the road section at a higher speed. The reason is that our method can complete the lane change in time to avoid entering the following state when the speed of the front vehicle is slow. The contrast method is unable to complete the lane change and is in the state of following for a long time, causing the speed to be restricted by the preceding car.
At about 80 seconds, the use of the contrast method caused severe fluctuations in speed. The vehicle is making frequent decision-making jumps at this time, which corresponds to the 2000m in Figure 13. This seriously affects the comfort of the ride, and there are great safety hazards, which may cause traffic accidents.

\subsection{Real-time analysis of different traffic flow densities}

\begin{figure}[htbp] 
	\centering
	\includegraphics[height=6cm,width=8.5cm,angle=0,scale=1.0]{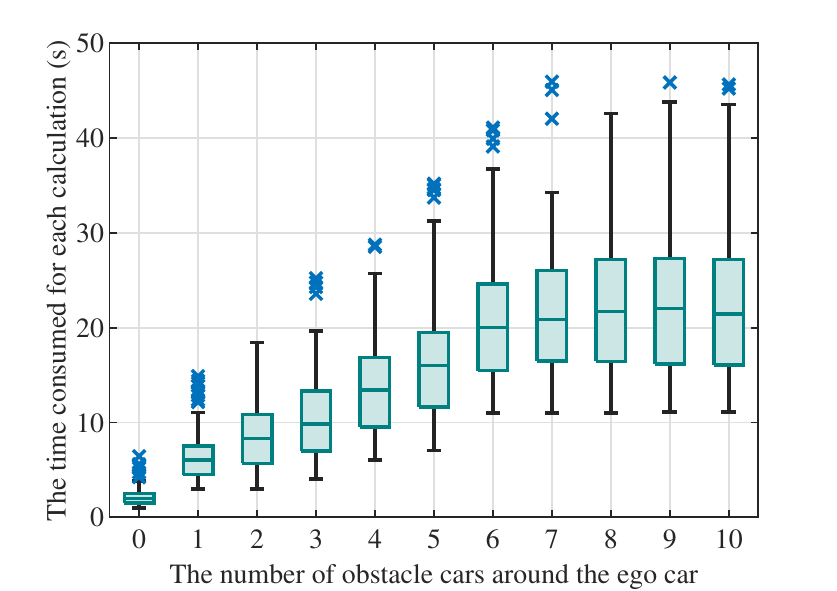} 
	\caption{ Illustration of the relationship between the number of surrounding obstacles and time-consuming. } 
	\label{fig:15} 
\end{figure}

Real-time performance is a very important indicator for evaluating motion planning algorithms. We analyzed the impact of the number of obstacles on the solution time within the perception range, from 300 meters in front to 100 meters behind. It can be seen from Figure 15 that as the number of surrounding obstacles increases, the time-consuming of each optimization cycle is increasing. When the number of obstacles is greater than 6, time-consuming tends to stabilize. The reason is that the number of estimated obstacle states in the sensing range is saturated, resulting in the number of collision avoidance constraints almost unchanged. In general, our method can complete the solution within 50 ms, making the motion planner run at a frequency of 20+HZ.

We use the previous solution as the initial value for the next iteration. It is a simple but very effective optimization acceleration method. In most cases, the state variables solved at adjacent moments will not change suddenly, because the vehicle has a great inertia. Therefore, such a warm start method can accelerate the convergence of the optimization.

\section{Conclusion}
In this paper, we propose an autonomous vehicle motion planning method in dense dynamic scenarios. The main contribution is to propose an instantaneous analysis model to analyze the collision relationship of obstacles in the spatio-temporal domain to remove redundant constraints and reduce computational complexity. Through analysis of typical scenes and experiments in dense traffic flows, this method can plan a safe and comfortable trajectory, avoid decision jumps in dense dynamic scenes, and improve traffic efficiency. In the future work, we will transform this method to the Frenet coordinate system. And deploy it in real vehicles to test on real highways.

\bibliographystyle{IEEEtran}
\bibliography{ref}
\end{document}